\documentclass[conference]{IEEEtran}
\IEEEoverridecommandlockouts
\usepackage{cite}
\usepackage{amsmath,amssymb,amsfonts}
\usepackage{graphicx}
\usepackage{tabularx}
\usepackage{textcomp}
\usepackage{xcolor}
\def\BibTeX{{\rm B\kern-.05em{\sc i\kern-.025em b}\kern-.08em
    T\kern-.1667em\lower.7ex\hbox{E}\kern-.125emX}}
\usepackage{float}
\usepackage{amsfonts}
\usepackage{subcaption}
\usepackage{caption}
\usepackage{booktabs}
{}
{}
{}
\usepackage{tikz}
\usetikzlibrary{shapes, arrows}
\usepackage{pdfpages}
\usepackage{pgfplots}
\usetikzlibrary{arrows,automata}
\usepackage[latin1]{inputenc}
\usepackage{multirow}
\usepackage{blindtext}
\usepackage{lipsum}
\usepackage{pgfplots}
\usepackage{mathtools}

\usepackage{balance}
\usepackage{comment}
\usepackage{amsmath}
\usepackage{graphics}
\usepackage{pgfplots,tikz}
\usepackage{xcolor}
\usepackage{pagecolor}
\usetikzlibrary{decorations, decorations.text,backgrounds}
\colorlet{NextBlue}{red!25!green!50!blue!75}
\usepackage{algorithm,algpseudocode}
\usepackage[acronym]{glossaries}
\makeglossaries
\usepackage{verbatim}
\usetikzlibrary{shapes}
\tikzstyle{line} = [draw,-latex']

\begin{document}

\title{CNN Autoencoder Resizer: A Power-Efficient LoS/NLoS Detector in MIMO-enabled UAV Networks \\

\thanks{}
}

\author{\IEEEauthorblockN{1\textsuperscript{st} Azim Akhtarshenas}
\IEEEauthorblockA{\textit{dept. of Telecommunication Engineering} \\
\textit{Universitat Polit\`ecnica de Val\`encia}\\
Valencia, Spain\\
aakhtar@doctor.upv.es}
\and
\IEEEauthorblockN{2\textsuperscript{nd} Navid Ayoobi}
\IEEEauthorblockA{\textit{dept. of Computer Science} \\
\textit{University of Houston}\\
Houston, USA \\
nayoobi@cougarnet.uh.edu}
\and
\IEEEauthorblockN{3\textsuperscript{rd} David Lopez-Perez}
\IEEEauthorblockA{\textit{dept. of Telecommunication Engineering} \\
\textit{Universitat Polit\`ecnica de Val\`encia}\\
Valencia, Spain \\
D.lopez@iteam.upv.es}
\and
\IEEEauthorblockN{4\textsuperscript{th} Ramin Toosi}
\IEEEauthorblockA{\textit{dept. of Electrical and Computer Engineering} \\
\textit{University of Tehran}\\
Tehran, Iran\\
r.toosi@ut.ac.ir}
\and
\IEEEauthorblockN{5\textsuperscript{th} Matin Amoozadeh}
\IEEEauthorblockA{\textit{dept. of Computer Science} \\
\textit{University of Houston}\\
Houston, USA \\
mamoozad@cougarnet.uh.edu}
}

\maketitle

\begin{abstract}

Optimizing the design, performance, and resource efficiency of wireless networks (WNs) necessitates the ability to discern Line of Sight (LoS) and Non-Line of Sight (NLoS) scenarios across diverse applications and environments. 
Unmanned Aerial Vehicles (UAVs) exhibit significant potential in this regard due to their rapid mobility, aerial capabilities, and payload characteristics. 
Particularly, UAVs can serve as vital non-terrestrial base stations (NTBS) in the event of terrestrial base station (TBS) failures or downtime. 
In this paper, 
we propose CNN autoencoder resizer (CAR) as a framework that improves the accuracy of LoS/NLoS detection without demanding extra power consumption.  
Our proposed method increases the mean accuracy of detecting LoS/NLoS signals from $66\%$ to $86\%$, 
while maintaining consistent power consumption levels.
In addition, 
the resolution provided by CAR shows that it can be employed as a preprocessing tool in other methods to enhance the quality of signals.
\end{abstract}


\section{Introduction}

Line of sight (LoS) occurs when there is no physical blockage in the direct path among the transmitter and the receiver, 
while non-line-of-sight (NLoS) takes place when there is no direct view from the transmitter to the receiver due to obstacles, 
and the signal arrives to receiver through different paths than the LoS one,
e.g. reflection, refraction. 
NLoS results in time-arrival differences and low tracking accuracy. 
Therefore, discriminating between LoS and NLoS propagation and correcting NLoS errors have important ramifications in designing and developing wireless networks. 
In particular, 
LoS and NLoS information can be used by operators and other stakeholders providing location-based services to enhance their services. 
Using traditional approaches \cite{wang2023survey, xiao2014non,yarkan2006identification,fan2019non,xiao2013identification} as well as Artificial Intelligence-based (AI) algorithms \cite{wei2022nlos, nguyen2018nlos, zhou2015wifi}, 
researchers and scholars have addressed and suppressed the errors caused by NLoS propagation to a certain extent.

Unmanned Aerial Vehicles (UAVs) have recently garnered significant interest as a robust tool to accelerate the process of identifying LoS/NLoS scenarios.
UAVs are inherently mobile platforms, 
capable of navigating through the three-dimensional space.
They are also easy to deploy due to their compact size, lightweight construction, and ease of transport. 
Moreover, UAVs may be equipped with communication systems that enable real-time data transmission between the vehicle and ground users. 
Considering all these aspects, 
UAVs can be quite effective in determining whether propagation is LoS or NLoS, 
and minimizing issues arising due to NLoS propagation \cite{scazzoli2020deep, yuan2022uav, vidan2023ultra, tian2019cost, ding2017uplink}.
To further enhance the capability to discriminate between LoS and NLoS signals, 
one common approach is to increase the number of receivers deployed on the receiver side,
e.g. the UAV. 
However, while this method enhances the accuracy of detection, 
it also raises power consumption concerns, 
as UAVs are already power-hungry devices \cite{mohsan2023unmanned}.


To address this challenge, 
and better discriminate among LoS and NLoS propagation in UAV use cases,
we introduce a novel CNN autoencoder resizer (CAR) framework in this paper. A CNN autoencoder is a type of autoencoder that uses CNNs as its building blocks. It combines the principles of CNNs and autoencoders to efficiently encode and decode image data, leveraging the spatial hierarchies learned by convolutional layers \cite{berahmand2024autoencoders, akhtarshenas2022open}. Regarding this, CAR concurrently trains two autoencoders, 
enabling the mapping of a low-dimensional channel output to a high-dimensional counterpart, 
thereby augmenting signal resolution. 
Our findings demonstrate that integrating CAR into the signal processing pipeline, 
followed by an input to a binary classifier, 
significantly enhances accuracy of detecting LoS and NLoS signals without necessitating additional power consumption.
In other words, 
CAR allows to learn how to discriminate between LoS and NLoS using measurements over lager receiving antenna arrays, 
and reuses that knowledge when queried with measurements from smaller receiving antenna arrays,
thus keeping energy consumption at bay. 

\subsection{Related Works}

Wang et. al, studied efficient LoS/NLoS identification and suppression methods in \cite{wang2023survey, xiao2014non}. 
Throughout their research endeavor, 
these researchers proposed the integration of an artificial intelligence (AI)-based methodology and a geometric property evaluation technique to identify NLoS propagation errors. 
Moreover, they highlighted the potency of weighting-based, constrained-based and path tracing-based optimization techniques in effectively mitigating the impact of these errors. 
The authors in \cite{xiao2013identification, yarkan2006identification} developed two methods for identifying NLoS situations and lessening their consequences. 
These methods solely relied on received signal strength (RSS) measurements, 
and used both hypothesis testing and machine learning (ML) techniques to craft the respective optimization problems and solvers through which NLoS conditions were identified. 
These frameworks considered various existing additional information and user demands. 
A computationally simple numerical method for coherent receivers was presented in \cite{fan2019non}, 
which enabled the discrimination between the LoS and NLoS stages of a transmission based on the autocorrelation features of individual received pathways.
In \cite{nguyen2018nlos}, 
the proposed ultra-wideband (UWB) positioning system integrated an effective NLoS recognition strategy, 
which utilized a multiple input learning (MIL) neural network model with channel impulse response (CIR) and time-frequency diagram of CIR (TFDOCIR).
With the objective of categorizing LoS and NLoS channels, 
Zimu et al. in \cite{zhou2015wifi} presented a combined convolutional neural network (CNN) and recurrent neural network (RNN) design. 
Using RNNs, 
they extracted features from the time-varying characteristics of both their received signal strength and channel state information measurements, 
while, 
in a second phase, 
they used CNNs to extract features from the frequency-domain characteristics of their channel state information measurements.
These features where later used to classify LoS and NLoS channels. 
To distinguish between LoS and NLoS components, 
the authors in \cite{wei2022nlos} used an unsupervised machine learning (UML) technique based on Gaussian mixture models. 
The main benefit of using UML is that it does not need databases to be explicitly and rigorously labeled at a specific location,
as in previous methods. 

The use of UAVs for propagation detection has received a lot of interest recently. 
For user equipment (UE), 
which are attempting to connect to an aerial base stations (ABS) via the physical random access channel (PRACH), 
Davide et al. in \cite{scazzoli2020deep} presented a scheme to determine LoS and NLoS conditions using CNNs, 
and evaluated its accuracy under different heights of the UAV. 
The authors in \cite{yuan2022uav} developed a method for UAV search and localization,
which can identify an unspecified number of victims. 
This approach built on effectively minimizing NLoS localization errors by utilizing the UAV as a mobile anchor for measuring time of arrival (TOA). 
The method, 
consisting of trajectory planning and localization, 
chooses waypoints to devise an appropriate search trajectory, 
wherein the UAV can collect a large amount of measurement data across the deployment area.
In this way, 
at least three anchors for localization are guaranteed for each target, 
even if it is in NLoS. 
In the localization phase, 
a maximum likelihood estimator was employed to predict both the target locations and the average NLoS bias. 
The authors in \cite{vidan2023ultra} aimed to test and show the viability of an UWB indoor GPS system for use in rescue and search operations in GNSS-denied areas,
like the interior of constructions damaged by natural and human-made disasters. 
Under LoS and NLoS scenarios, 
they mainly evaluated the reliability and safety of an ad hoc UWB network through indoor navigation for UAVs.
%
Using inexpensive COTS machines, 
the scientists in \cite{tian2019cost} presented a unique RSS-based NLoS UAV identification technique. 
In stark contrast to current studies, 
their approach was not dependent on UAVs being at LoS locations,
and did not require any manually imposed fly patterns.
Instead, it utilized the inherent flying routines of UAVs. 
To achieve their objective, 
the authors used deep learning to extract collective properties from incoming radio frequency (RF) signals, 
like those found in previous research. 
They began by categorizing any identified RF signals (i.e., their RSS observations) into LoS and NLoS signals without considering the presence of UAVs. 
They finally used a Long-Short-Term-Memory structure to recognize UAVs on NLoS signals, 
accordingly.

Using a piecewise linear path loss approach that incorporates both LoS and NLoS propagations, 
Ding et al. in \cite{ding2017uplink} examined the area spectral efficiency (ASE) and coverage probability in the uplink (UL) of dense small cell networks (SCNs). Comparing with earlier studies that did not distinguish between LoS and NLoS transmissions, 
these studies demonstrated that the effectiveness of LoS and NLoS transmissions in the ASE of the UL of dense SCNs is substantial, 
both statistically and qualitatively. 
This shows the importance of characterizing LoS and NLoS propagation in network planning too. 
In general, 
poor LoS/NLoS identification performance leads to inefficient network design and implementation.
One straightforward solution for addressing it and improving performance is increasing the number of transmitters or receivers with various distances, as this allows for an increase in the resolution with which the channel is assessed \cite{sukhanov2021manipulating, huang2020machine}. 

This, however, will bring about more complexity and computation,
and in turn an extra power consumption, 
which pose a serious problem for the already power-hungry telecom gear and UAVs. 

\begin{figure*}[!t]
    \centering
    \includegraphics[width=0.95\textwidth]{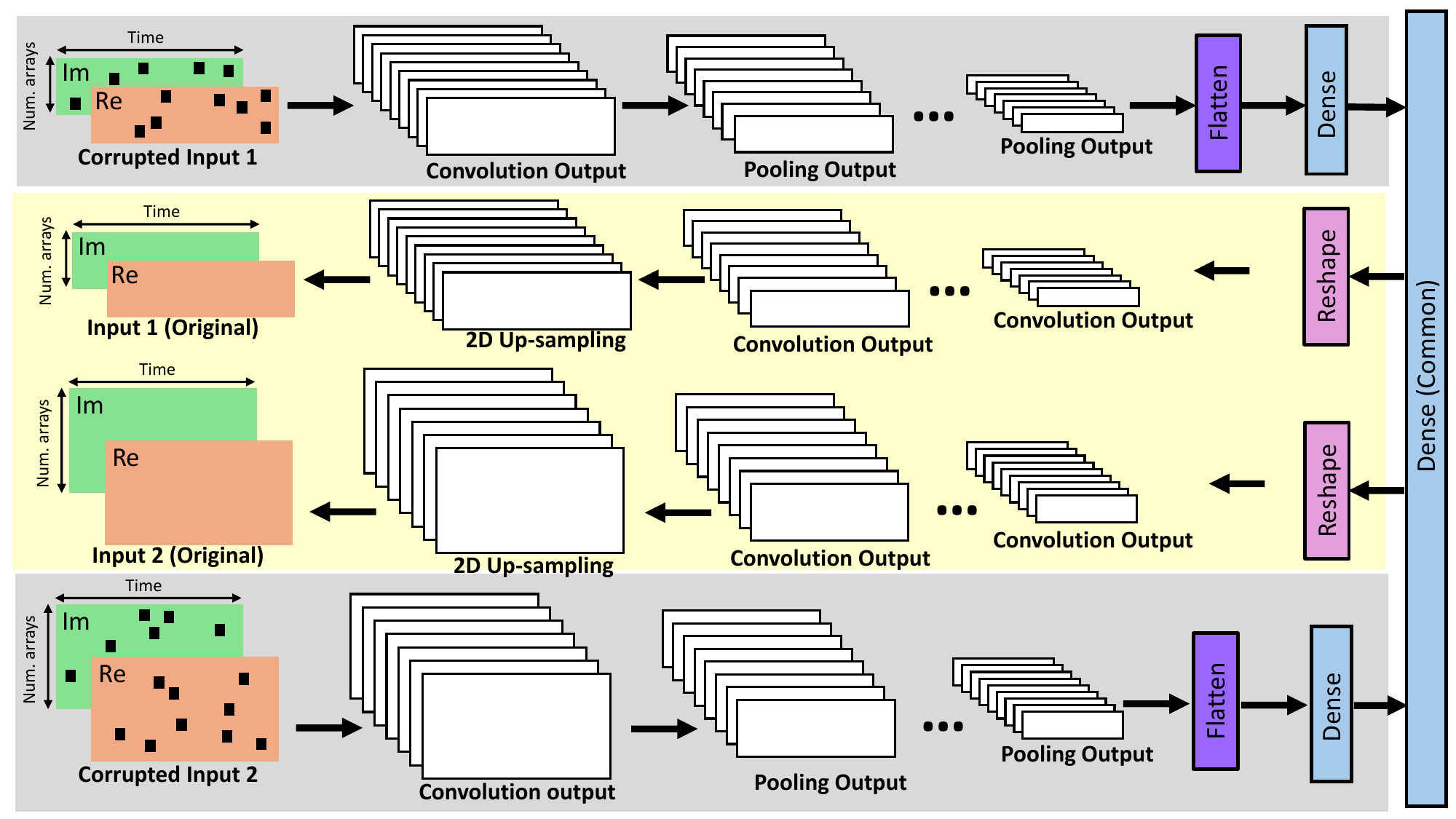}
    \caption{The architecture of the proposed CNN autoencoder resizer (CAR). CAR includes two autoencoders that share a common latent vector. The first autoencoder processes the lower dimensional channel outputs while the second one processes the higher dimensional channel outputs. After training stage, CAR is able to map a low-dimensional channel output to a latent vector using the encoder in the first autoencoder, and then decodes it to a higher dimensional channel output using the decoder of the second autoencoder. The encoders and decoders are highlighted by gray and yellow backgrounds, respectively.}
    \label{fig:autoencoder}
\end{figure*}

\subsection{Motivation and Contribution}

Despite the abundance of studies on LoS/NLoS signal identification, 
existing approaches tend to prioritize accuracy at the expense of increased power consumption and extra complexity due to the more hardware deployed to perform the assessment. 
To simultaneously address these issues, 
and mitigate potential NLoS propagation errors, 
we present CAR in this paper. 
Our key contributions are delineated as follows:
 \begin{itemize}
     \item CAR is introduced as an tool for refining the resolution of received channel outputs.
     This framework can thus serve as a preprocessing step in various methods to enhance the quality of the signals at hand.
     \item The proposed CAR architecture is optimized for enhancing the accuracy of LoS and NLoS signal discrimination at UAVs,
     while maintaining consistent power consumption levels.
 \end{itemize}
In short, 
and to provide physical intuition, 
CAR trains the system using receiving antenna arrays of various sizes, 
and allows to reuse the knowledge extracted from larger arrays in inferences performed with data of smaller arrays. 
In this way, classification accuracy is enhanced,
while allowing to work with smaller antenna arrays and thus radios that consume less power. 


\section{System model} 

When terrestrial networks are out of service or collapse, 
UAVs, 
as non-terrestrial new base stations, 
may come to stage and serve UEs,
keeping the communication services running, 
even under lower signal to noise ratios (SNRs). 
In these cases, 
as in any standard cellular network,
the PRACH, 
and the related specified initial access protocols, 
enable UEs (like smart devices, mobile phones or internet of things devices) to associate to the base stations mounted on UAVs. 
To obtain information about LoS/NLoS propagation characteristics, 
we are going to use the power delay profile (PDP) of the signals transmitted in the PRACH.

It should be noted that the specified preambles used in the PRACH of a cellular network are based on Zadoff-Chu (ZC) sequences, 
which can be found at \cite{TS}.
A ZC sequence of length $K$ is given by:
\begin{equation}   
    zc_{m}(k) = e^{-j\frac{\pi m k (k+1)}{K}},
\end{equation}
where $m$ is the root index (an integer relatively prime to $K$) of the particular ZC sequence, 
$k$ is the sequence index, 
ranging from 0 to $K-1$, 
and $K$ is the length of the sequence.

Let us assume that $s(i)$ denotes the ZC sequence signal received by the base station mounted on a UAV, 
its PDP can be represented as follows \cite{scazzoli2020deep}:
\begin{equation}   
    PDP =\left|{\sum_{i=1}^{K-1}s(i)zc_{m}^{*}(i+1){\rm MOD}(K)}\right|^2.
\end{equation}
After detecting the ZC sequence signal at the receiver, 
our base station mounted on the UAV needs to identify the type of propagation in terms of LoS and NLoS. 
To this end, 
we propose the use of the the root mean square (RMS) delay spread of the PDP, 
which measures the dispersion of the multi-path components in a communication channel under various propagation conditions. 
The RMS delay spread is small in LoS conditions, 
as the delay between meaningful multi-path components is small. 
In NLoS conditions, 
the RMS delay spread is larger due to the more diverse paths that the multi-path components encounter.  
Therefore, 
we believe that RMS delay spread can be a good metric for LoS/NLoS identification \cite{huang2019angular}. 

Let us assume that $t_{i}$ and $p_{i}$ indicate the delay and power of the $i^{th}$ path, respectively.
The overall RMS-delay spread can be calculated as: 
\begin{equation}
    T_{\rm RMS} =\sqrt{\dfrac{\sum_{i=1}^{n}p_{i}(t_{i}-\Bar{t})^{2}}{\sum_{i=1}^{n}p_{i}}},
\end{equation}
where
\begin{equation}
    \Bar{t} = \sqrt{\dfrac{\sum_{i=1}^{n}p_{i}t_{i}}{\sum_{i=1}^{n}p_{i}}}.
\end{equation}
Note that we assume that receivers use the above formulation to characterise its corresponding RMS delay spread.

\section{Proposed Method and Architecture}

In the following, 
we elaborate on the data available and the proposed method to differentiate LoS from NLoS conditions.  

\begin{figure*}[t!]
    \centering
    \includegraphics[width=\textwidth]{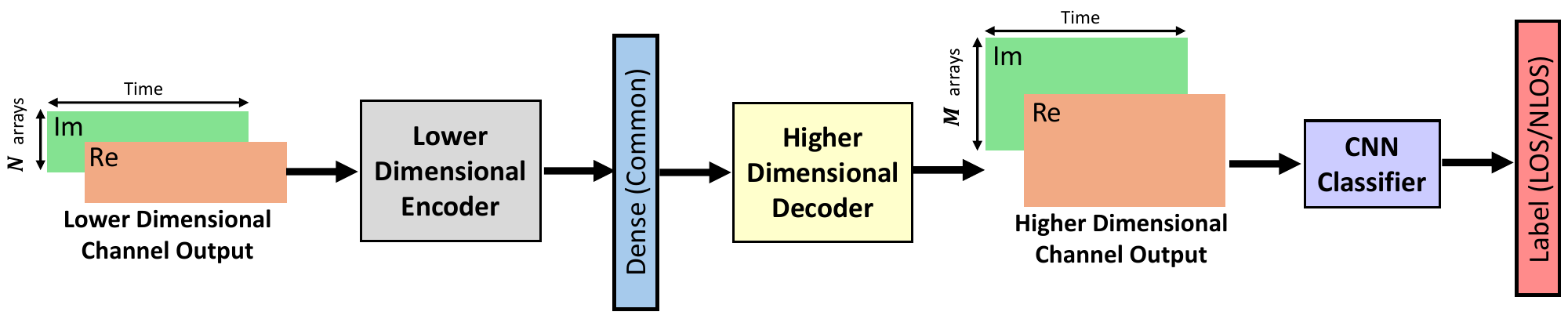}
    \caption{The pipeline of labeling channel outputs. The low-dimensional channel output is fed to the first encoder. The second decoder uses encoded vector and generates its high-dimensional counterpart. The binary CNN classifiers then processes this high-dimensional channel output to predict its label.}
    \label{fig:classifier}
\end{figure*}

\subsection{Dataset}

The dataset generation process closely resembled the approach outlined in reference \cite{scazzoli2020deep}. 
We synthesized our dataset within a 3D urban context, 
occupying a total area of $50\,m \times 50\,m $. 
The UAVs, 
equipped with an $N \times N$ antenna array, 
were positioned in accordance with a normal distribution having a mean of zero and a standard deviation of $5\,m$, 
centered within the area at an altitude of $75\,m$. 
Five distinct UAV locations were chosen randomly, 
and for each location, 
$2000$ ground UEs were dispersed randomly within the designated area. 
The PRACH signals emitted by ground UEs were generated using the long term evolution (LTE) PRACH tool available in the MATLAB toolkit. 
Subsequently, 
we computed the channel response by assuming adherence to the QuaDRiGa channel model \cite{jaeckel2014quadriga}. 
Our dataset was curated to ensure an equal split of samples between LoS and NLoS scenarios. 
The resulting channel output is organized as a tensor with dimensions $(N^2, T, 2)$, 
where $N^2$ corresponds to the total number of receive antennas, 
and $T$ signifies the duration of the received signal. 
The last dimension of the tensor represents the real and imaginary parts of the channel output. 


\subsection{CNN Autoencoder Resizer (CAR)}

Fig. \ref{fig:autoencoder} depicts the architecture of the proposed CNN autoencoder Resizer (CAR). 
CAR comprises two CNN autoencoders that share a common latent vector produced by the ``Dense (Common)'' block shown in the figure. 
The first encoder is designed to accommodate a lower-dimensional channel output as its input. 
This input is subject to corruption by randomly setting a subset of its elements to zero, 
a process influenced by the corruption factor hyperparameter. 
The corrupted input proceeds through multiple layers of convolutional and mean pooling operations before being transformed into a vector via a flattening layer and a dense layer. 
The first decoder then accepts the latent vector, 
reshapes it into a tensor, 
and navigates through multiple convolutional layers and 2D up-sampling layers to rectify the corrupted input and reconstruct the original channel output.
The second autoencoder operates analogously to the first autoencoder, 
albeit dealing with the higher-dimensional channel output produced by a UAV with a larger array configuration.
Recall that in our setup, we have measurements taken with UAVs with different antenna array capabilities. 


Leveraging a shared latent vector across its two autoencoders, 
CAR effectively encodes low-dimensional channel outputs with its first encoder. 
Subsequently, 
it employs the second decoder to decode the latent vector, 
yielding a higher-dimensional channel output. 
This pipeline is depicted in Fig. \ref{fig:classifier}. 
The proposed method enhances the resolution of the channel output, 
thereby helping to improve the performance of the subsequent LoS and NLoS classification task when using smaller antenna arrays, 
while maintaining power efficiency,
as it will be shown in the results section of this paper. 


\subsection{Binary CNN Classifier}

For the discrimination LoS from NLoS signals, 
we utilize a binary CNN classifier. 
The classifier operates on the high-dimensional channel output derived from CAR, 
undergoing processing across its convolutional layers, mean pooling layers, and two dense layers. 
By incorporating a sigmoid activation function in its final layer, 
the classifier aims to predict the label assigned to the primary low-dimensional channel output.


\section{Result and discussion}

In the following, 
we describe the experimental evaluation used to describe the goodness of the proposed method and discuss the results. 

\subsection{Experimental setup}

Experiments were carried out utilizing a single GPU, 
specifically the NVIDIA GeForce RTX 3090 equipped with 25GB memory. 
To mitigate overfitting, 
hyperparameters underwent fine-tuning through 5-fold cross-validation. 
The learning rate, batch size, and the number of epochs for the autoencoder and for the classifier were set at $0.001$, $32$, $2$, and $15$, respectively. 
The Adam optimizer was employed for training CAR, 
while the binary classifiers were trained using the stochastic gradient descent (SGD) optimizer.
Mean squared error was utilized as the loss function for updating CAR weights, 
while binary cross-entropy was chosen for learning the weights of the binary classifiers. 
We divided our dataset into an $80/20$ ratio for training and testing sets.

The required parameters for PRACH and Quadriga Urban D2D channels were assigned as those in \cite{scazzoli2020deep} and \cite{jaeckel2014quadriga}, respectively.

\subsection{Comparison of mapped channel output with the original and equivalent higher channel output}

An experiment was conducted to assess the efficiency of the proposed CAR in improving the accuracy of distinguishing between LoS and NLoS signals. 
Three distinct binary classifiers were trained using the low-dimensional channel output, the high-dimensional mapped channel output, and the equivalent original high-dimensional channel output. 
The low-dimensional channel output was generated by an array of size $4 \times 4$, 
while the equivalent high-dimensional channel output was produced by an array of size $16 \times 16$. 
To obtain the equivalent original high-dimensional channel output, 
the same information as the low-dimensional channel output was used, 
with the exception of the number of arrays. 

\begin{figure}[!t]
    \centering
    \includegraphics[width=\linewidth]{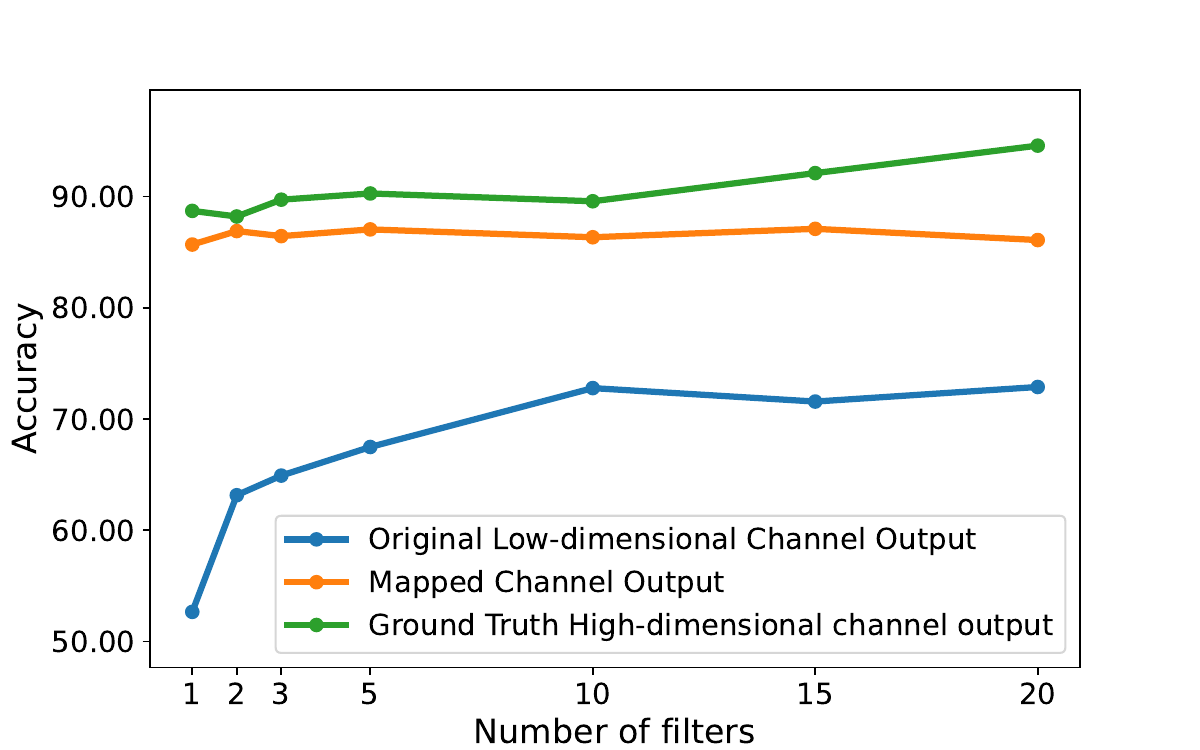}
    \caption{The performance of CAR on the accuracy of detection of LoS/NLoS by mapping the low-dimensional channel output to high-dimensional channel output.}
    \label{fig:accuracy}
\end{figure}

We incremented the number of convolutional and mean pooling filters used in CNN autoencoder resizer and binary classifiers from $1$ to $20$, 
and measured the accuracy of the classification using the testing set. 
Fig. \ref{fig:accuracy} shows the results,
which indicate that the mapped channel output outperforms the binary classifier trained on the low-dimensional channel output for all filter numbers.
It increases the accuracy of detecting LoS and NLoS signals from $66\%$ to $86\%$ in average.
Although the mapped channel output falls short of achieving the accuracy of the original high-dimensional channel output, 
it approaches the performance of the high-dimensional channel output, 
while remaining significantly more power-efficient. 

In other words, 
in the proposed CAR method, 
arrays of size $4 \times 4$ and $16 \times 16$ are only required during the training phase to develop a function that maps $4 \times 4$ array data to $16 \times 16$ array data. 
Once the training is complete, 
the architecture dispenses with the first decoder and the second encoder. 
Thus, in the testing phase, only data from the $4 \times 4$ array is collected and fed into the first encoder. 
This encoder converts the low-dimensional channel output into a latent vector, 
which is then processed by the second decoder to generate a high-dimensional channel output. 
This method allows for the production of high-resolution channel outputs using only the low-dimensional array, 
without the need for collecting the data from a high-dimensional array in practical use. 
This preserves the low-dimensional array's power efficiency while enhancing the resolution of the channel output and hence, 
improving detection accuracy.


\subsection{Evaluating the effect of the length of the filters along time access}

We assessed the impact of varying the length of convolutional and mean pooling filters along the time dimension to capture temporal patterns in channel output. 
The bar plot illustrated in Fig. 4 indicates that CAR exhibits robust behavior across different filter lengths along the temporal axis, 
whereas for unmapped signals, 
this parameter necessitates tuning.

\begin{figure}
    \centering
    \includegraphics[width=\linewidth]{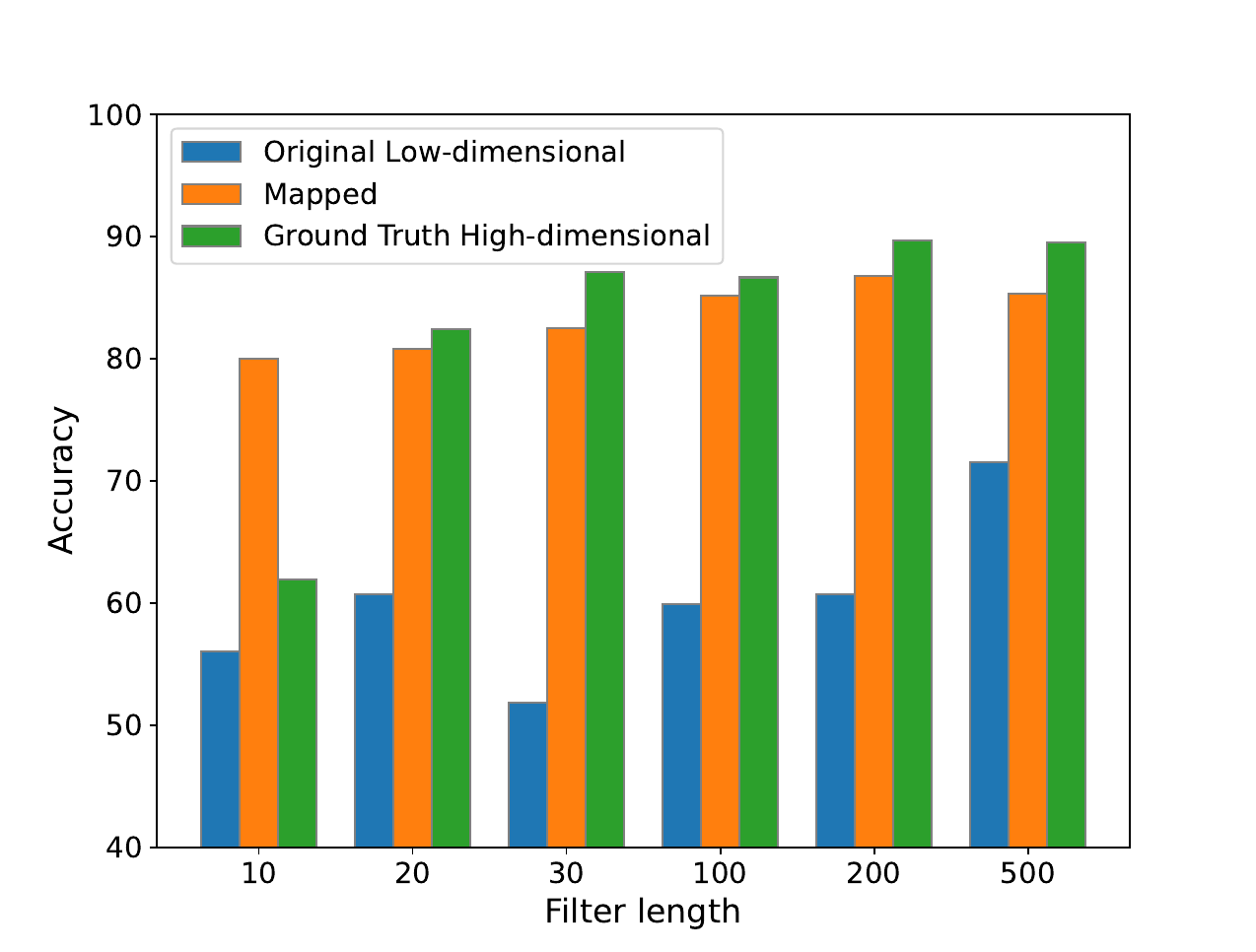}
    \caption{The effect of changing the length of filters along time axis in terms of accuracy.}
    \label{fig:enter-label}
\end{figure}


\section{Summary and Conclusion}

The presence of NLoS signals introduces temporal disparities and reduces tracking accuracy. 
The ability to differentiate between LoS and NLoS propagation is essential for the optimization of wireless network architectures. 
In this study, 
we presented the CNN autoencoder resizer (CAR) framework, 
which trains two autoencoders simultaneously to map low-dimensional channel outputs to higher-dimensional representations, 
thereby improving signal resolution. 
Our findings indicate that the integration of CAR into the signal processing pipeline substantially improves the accuracy of LoS/NLoS detection without imposing additional power consumption.

\balance

\section*{Acknowledgement}

This research is supported by the Generalitat Valenciana
through the CIDEGENT PlaGenT, Grant CIDEXG/2022/17,
Project iTENTE, and by the action CNS2023-144333, financed
by MCIN/AEI/10.13039/501100011033 and the European
Union ``NextGenerationEU''/PRTR.

\bibliographystyle{IEEEtran}
\bibliography{references.bib}
\vspace{12pt}

\end{document}